%% file: root.tex
\documentclass[letterpaper, 10 pt, conference]{ieeeconf}  

\IEEEoverridecommandlockouts                              
\usepackage{mathtools,amssymb,commath}
\usepackage{amsmath}
\usepackage{graphicx,import}
\usepackage{verbatim}
\usepackage{color}
\usepackage{hyperref}
\hypersetup{colorlinks=true}

\DeclareMathOperator{\real}{Re}

\usepackage{mathtools,amsthm}

\title{\LARGE \bf Exploiting Friction in Torque Controlled Humanoid Robots
}

\author{Gabriele Nava, Diego Ferigo and Daniele Pucci $^{1}$
\thanks{*This project has received funding from the European Union’s Horizon
2020 research and innovation programme under grant agreement No. 731540
(An.Dy). The content of this publication is the sole responsibility of the
authors. The European Commission or its services cannot be held responsible
for any use that may be made of the information it contains.
}
\thanks{$^{1}$ All authors belong to the Italian Institute of Technology,
        Via Morego 30, Genoa, Italy
        {\tt\small name.surname@iit.it}}%
}

\include{definitions}

\begin{document}

\maketitle
\thispagestyle{empty}
\pagestyle{empty}

\begin{abstract}
A common architecture for torque controlled humanoid robots consists in two nested loops. The outer loop generates desired joint/motor torques, and the inner loop stabilizes these desired values. In doing so, the inner loop usually compensates for joint friction phenomena, thus removing their inherent stabilizing property that may be also beneficial for high level control objectives. This paper shows how to exploit friction for joint and task space control of humanoid robots. Experiments are carried out on the humanoid robot iCub.

\end{abstract}

\import{tex/}{intro}
\import{tex/}{background}
\import{tex/}{method}

\import{tex/}{results}

\import{tex/}{conclusions}

\addtolength{\textheight}{0cm}     

\bibliographystyle{IEEEtran}
\bibliography{IEEEabrv,Biblio}

\end{document}

%% file: tex/definitions.tex
\makeatletter
\newsavebox\myboxA
\newsavebox\myboxB
\newlength\mylenA
\newcommand*\xoverline[2][0.75]{%
    \sbox{\myboxA}{$\m@th#2$}%
    \setbox\myboxB\null
    \ht\myboxB=\ht\myboxA%
    \dp\myboxB=\dp\myboxA%
    \wd\myboxB=#1\wd\myboxA
    \sbox\myboxB{$\m@th\overline{\copy\myboxB}$}
    \setlength\mylenA{\the\wd\myboxA}
    \addtolength\mylenA{-\the\wd\myboxB}%
    \ifdim\wd\myboxB<\wd\myboxA%
       \rlap{\hskip 0.5\mylenA\usebox\myboxB}{\usebox\myboxA}%
    \else
        \hskip -0.5\mylenA\rlap{\usebox\myboxA}{\hskip 0.5\mylenA\usebox\myboxB}%
    \fi}
\makeatother

%% file: tex/intro.tex
\section{INTRODUCTION}
\label{sec:introduction}

A humanoid robot is usually required to operate out of a protected and well-known workspace and to physically interact with a dynamic, human-centered environment. In this context, the robot is required to balance, perform manipulation tasks and -- even more important -- to safely interact with humans. 
The importance of controlling the robot interaction with the environment calls for the design of torque and impedance control algorithms, capable of exploiting the forces the robot exerts at contact locations for performing dynamic tasks \cite{ott2011,farnioli2015}.
However, despite decades of research in the subject, torque controlled humanoid robots are still a challenge for the robotics community. The variability of sensor load during locomotion, the inaccuracy of the force/torque sensing technology, and the nonlinearity of joint friction effects are only a few complexities impairing efficient robot torque control. Then, the importance of conceiving control laws ensuring a degree of robustness against some of these factors goes without saying. This paper contributes towards this direction by proposing  modifications of state-of-the-art control laws that allow them to exploit the inherent stabilizing nature of joint friction. The effect of these modifications is a system degree of robustness against poor joint velocity measurements, and an improvement of the tracking performances of the controlled system .

Similar solutions that try to exploit the \emph{natural dynamics} of the system for improving performances and energy efficiency have been proposed in literature, e.g. for robot walking \cite{Pratt99} or running \cite{Iida2005}. In particular, the  effect of friction at all stages of the robot mechanisms and between the robot and the environment plays an important role for the stability of the controlled system \cite{Miura2008,PANTELEY1998}. Previous works already considered the possibility of exploiting the friction exerted between the robot and the environment for controlling a robotic crawler \cite{Noselli2014}, and for the locomotion of a hopping robot \cite{Iida2005}. More generally, the \emph{passivity-based} control strategies try to exploit the passivity properties of the overall system for regulation tasks \cite{li2012}, and can also be extended for addressing tracking problems \cite{schaffer2007}. 

When dealing with humanoid robots, the fixed-base assumption may be a limitation for tasks such as walking. An alternative solution is to make use the \textit{floating base} formalism \cite{Featherstone2007}, i.e. none of the robot link is assumed to have an \textit{a priori} constant pose w.r.t. an inertial reference frame. In this case, the control problem is further complicated by the system's underactuation, since it forbids full state feedback linearization ~\cite{Acosta05}. 

An effective technique for controlling floating base robots with rigid joints is the operational space formulation, where the control objective is often the stabilisation of the robot \emph{centroidal momentum}~\cite{Orin2013}. The controllers designed for this objective are usually referred to as 
\textit{momentum-based} controllers~\cite{Lee2012}. 
Momentum control can be achieved by controlling the forces the robot exerts at contact locations \cite{Stephens2010,Herzog2014,Frontiers2015}, and these forces are then generated
by the robot joint torques.
To get rid of the (eventual) actuation redundancy associated with momentum control, a lower priority task is usually added during the stabilisation of the robot momentum. 
This secondary task aims at imposing a desired joint robot configuration, and plays a pivotal role for the stabilization of the system \textit{zero dynamics}~\cite{nava2016}. 

The aim of this paper is the development of a torque control framework for a humanoid robot that exploits the joint viscous and Coulomb friction to increase the system's robustness against noisy velocity measurements, and to improve the tracking of a desired reference trajectory. We first develop a control algorithm for fixed base robots that exploits friction for improving the tracking of a desired joint space trajectory, Then, we extend our formulation to the control of a floating base robot. In particular, the joints friction is exploited in order to improve the tracking of a desired momentum trajectory.

This paper is organized as follows. Section \ref{sec:background} recalls notation, system modelling, and the momentum based control strategy for balancing developed in our previous work \cite{nava2016}. In Section \ref{sec:method}, the modification of the control framework for exploiting joints friction is detailed. Experimental results on humanoid robot iCub~\cite{Nataleeaaq1026} are presented in Section \ref{sec:results}. 
Conclusions and perspectives conclude the paper.

%% file: tex/background.tex
\section{BACKGROUND}
\label{sec:background}

\subsection{Notation}

\begin{itemize}
\item $\mathcal{I}$ denotes an inertial frame of reference, with its $z$ axis pointing against the gravity. $\mathcal{B}$ denotes the \emph{base frame}, i.e. a frame attached to the robot \emph{base link}. The constant $g$ denotes the norm of the gravitational acceleration.
\item Given a matrix $A \in \mathbb{R}^{m \times n}$, we denote with $A^{\dagger}\in \mathbb{R}^{n \times m}$ its Moore Penrose pseudoinverse. 
\item $e_i \in  \mathbb{R}^m$ is the canonical vector, consisting of all zeros but the $i$-th component that is equal to one.
\item We denote with $m$ the total mass of the robot.
\end{itemize}

\subsection{Recalls on Robot Dynamics}

The robot is modelled as a multi-body system composed of $n + 1$ rigid bodies, called \emph{links}, connected by $n$ joints with one degree of freedom each. If no links have an \emph{a priori} constant position and orientation w.r.t. the inertial frame $\mathcal{I}$, the system is considered \emph{floating base}. 

The robot configuration space is the Lie group $\mathbb{Q} = \mathbb{R}^3 \times SO(3) \times \mathbb{R}^n$. An element $q \in \mathbb{Q}$ can be defined as the following triplet: $q = (\prescript{\mathcal{I}}{}p_{\mathcal{B}}, \prescript{\mathcal{I}}{}R_{\mathcal{B}}, s)$ where $\prescript{\mathcal{I}}{}p_{\mathcal{B}} \in \mathbb{R}^3$ denotes the position of the base frame $\mathcal{B}$ with respect to the inertial frame, $\prescript{\mathcal{I}}{}R_{\mathcal{B}} \in \mathbb{R}^{3\times3}$ is a rotation matrix representing the orientation of the base frame, and $s \in \mathbb{R}^n$ is the joint configuration. 

The velocity of the multi-body system can be characterized by the set $\mathbb{V} = \mathbb{R}^3 \times \mathbb{R}^3 \times \mathbb{R}^n$. An element of $\mathbb{V}$ is a triplet $\nu = (^\mathcal{I}\dot{p}_{\mathcal{B}},^\mathcal{I}\omega_{\mathcal{B}},\dot{s}) = (\text{v}_{\mathcal{B}}, \dot{s})$, where $^\mathcal{I}\omega_{\mathcal{B}}$ is the angular velocity of the base frame expressed w.r.t. the inertial frame, 
i.e. $^\mathcal{I}\dot{R}_{\mathcal{B}} = S(^\mathcal{I}\omega_{\mathcal{B}})^\mathcal{I}{R}_{\mathcal{B}}$. A more detailed description of the floating base formalism is provided in \cite{traversaro2017}.

We assume that the robot interacts with the environment by exchanging $n_c$ distinct wrenches. The equations of motion of the multi-body system can be described by applying the Euler-Poincar\'e formalism \cite[Ch. 13.5]{Marsden2010}:
\begin{align}
   \label{eq:system}
   {M}(q)\dot{{\nu}} + {C}(q, {\nu}){\nu} + {G}(q) =  B \tau + \sum_{k = 1}^{n_c} {J}^\top_{\mathcal{C}_k} f_k
\end{align}
where ${M} \in \mathbb{R}^{n+6 \times n+6}$ is the mass matrix, ${C} \in \mathbb{R}^{n+6 \times n+6}$ accounts for centrifugal and Coriolis effects, ${G} \in \mathbb{R}^{n+6}$ is the gravity vector, $B = (0_{n\times 6} , 1_n)^\top$ is a selector matrix, $\tau \in \mathbb{R}^{n}$ is a vector representing the joint torques, and $f_k \in \mathbb{R}^{6}$ denotes an external wrench applied by the environment to the link of the $k$-th contact. The Jacobian ${J}_{\mathcal{C}_k} = {J}_{\mathcal{C}_k}(q)$ is the map between the robot's velocity ${\nu}$ and the linear and angular velocity at the $k$-th contact link, i.e. $\text{v}_k = {J}_{\mathcal{C}_k}\nu$.

As described in \cite[Sec. 5]{traversaro2017}, it is possible to apply a coordinate transformation in the state space $(q,{\nu})$ that transforms the system dynamics~\eqref{eq:system} into a new form where the mass matrix is block diagonal, thus decoupling joint and base frame accelerations. Also, in this new set of coordinates,  the first six rows of Eq. \eqref{eq:system} are the so-called \emph{centroidal dynamics}\footnote{In the specialized literature, the term \emph{centroidal dynamics} is used to indicate the rate of change of the robot's momentum expressed at the center-of-mass, which then equals the summation of all external wrenches acting on the multi-body system \cite{Orin2013}.}. As an abuse of notation, we assume that system \eqref{eq:system} has been already transformed into this new set of coordinates, i.e.  \begin{IEEEeqnarray}{RCL}
    \label{centrTrans}
    M(q) &=& \begin{bmatrix} {M}_b(q) & 0_{6\times n} \\ 0_{n\times 6} & {M}_s(q) \end{bmatrix}, \quad
    H    = M_b \text{v}_{\mathcal{B}},
\end{IEEEeqnarray}
with ${M}_b \in \mathbb{R}^{6\times 6}, {M}_s \in \mathbb{R}^{n\times n}$, ${H}:=(H_L^\top,H^\top_\omega)^\top\in \mathbb{R}^6$  the robot centroidal momentum, and $H_L, H_\omega \in \mathbb{R}^3$  the linear and angular momentum at the center of mass, respectively. The base frame velocity $\text{v}_{\mathcal{B}} \in \mathbb{R}^6$ in the new coordinates yielding a block-diagonal mass matrix is given by $\text{v}_{\mathcal{B}} = (\dot{p}_c,\omega_o)$, where $\dot{p}_c \in \mathbb{R}^3$ is the velocity of the system's center of mass ${p}_c \in \mathbb{R}^3$, and $\omega_o \in \mathbb{R}^3$ is the so-called \emph{locked} (or average) angular velocity \cite{Orin2013}. When all the joint velocities are locked ($\dot{s} = 0$), $\omega_o$ represents the angular velocity of the robot, that now behaves as a single rigid body.
                           
Lastly, it is assumed that a set of holonomic constraints acts on System \eqref{eq:system}. These constraints may represent, for instance, a frame having a constant pose w.r.t. the inertial frame. In case this frame corresponds to the location at which a rigid contact occurs on a link, we represent the holonomic constraint as ${J}_{\mathcal{C}_k}(q) {\nu} = 0.$

Hence, the holonomic constraints associated with all the rigid contacts can be represented as
\begin{IEEEeqnarray}{RCL}
    \label{eqn:constraintsAll}
    {J}(q) {\nu} {=} 
    \begin{bmatrix}{J}_{\mathcal{C}_1}(q) \\ \cdots \\ {J}_{\mathcal{C}_{n_c}}(q)  \end{bmatrix}{\nu}  {=} 
    \begin{bmatrix} J_b & J_s  \end{bmatrix}{\nu}  
    &=& J_b {\text{v}}_{\mathcal{B}}+ J_s \dot{s} = 0,
    \IEEEeqnarraynumspace
\end{IEEEeqnarray}
 
\noindent with $J_b \in \mathbb{R}^{6n_c \times 6},J_s \in \mathbb{R}^{6n_c \times n} $. By differentiating the kinematic constraint \eqref{eqn:constraintsAll}, one obtains
\begin{equation}
    \label{eq:constraints_acc}
    J\dot{\nu}+\dot{J}\nu = J_b \dot{\text{v}}_{\mathcal{B}}+ J_s \ddot{s} + \dot{J}_b {\text{v}}_{\mathcal{B}}+ \dot{J}_s \dot{s} = 0.
\end{equation}

\subsection{Motors Dynamics}

The joints actuation is provided by $n$ electric brushless motors. We assume that motors and joints are rigidly connected to each other by means of the transmission element. The single joint rotation may be obtained by a linear combination of the actuators movements. The relationship between joint and motor positions is given by:
\begin{IEEEeqnarray}{LCL}
      \label{eq:jointMotorConstraint}
      s = \Gamma \theta 
  \end{IEEEeqnarray}
where $\theta \in \mathbb{R}^n$ are the motor positions and $\Gamma \in \mathbb{R}^{n \times n}$ is a matrix that accounts for the gear box ratios and for the coupling between the input and the output rotations of the coupling mechanism. Furthermore, we also make the following assumptions~\cite{schaffer2004,deluca1998}:

\begin{itemize}
    \item the friction of the mechanism is modelled as a combination of Coulomb and viscous friction only;
    \item the angular motor kinetic energy is due to its own spinning only, and the center of mass of each motor is along the motor axis of rotation.
\end{itemize}

\noindent The motors dynamics is then given by:
\begin{IEEEeqnarray}{LCL}
      \label{eq:motorsDynamics}
      I_m\ddot{\theta} +K_v\dot{\theta} +K_c \text{sign}(\dot{\theta}) & = & \tau_m -\Gamma^{\top}\tau,
\end{IEEEeqnarray}

\noindent with $I_m = \text{diag}(b_i) \in \mathbb{R}^{n \times n}, \  b_i > 0, \  i=1...n$ the motors inertia matrix, while $K_v = \text{diag}(k_{v_i}) \in \mathbb{R}^{n \times n}$ and $K_c = \text{diag}(k_{c_i}) \in \mathbb{R}^{n \times n}$ are diagonal matrices collecting all the viscous and Coulomb friction coefficients, respectively. $\tau_m \in \mathbb{R}^n$ are the motor torques.

In view of \eqref{eq:system}--\eqref{centrTrans}--\eqref{eq:motorsDynamics}, the equations representing the robot and motors dynamics are given by:
\begin{IEEEeqnarray}{LCL}
    \label{eq:systemFullCentroidal}
    M_b\dot{\text{v}}_\mathcal{B} + h_b  =   J_b^{\top}f     \IEEEyessubnumber \label{eq:floatingCentroidal} \\
    M_s\ddot{s} + h_s    =   J_s^{\top}f  + \tau              \IEEEyessubnumber \label{eq:jointsCentroidal}  \\
    I_m\ddot{\theta} +K_v\dot{\theta} +K_c \text{sign}(\dot{\theta}) & = & \tau_m -\Gamma^{\top}\tau  \IEEEyessubnumber
\end{IEEEeqnarray} 

\noindent subject to the contact constraints Eq. \eqref{eq:constraints_acc}. In particular, following Eq. \eqref{centrTrans} we partitioned Eq. \eqref{eq:system} into the floating base dynamics Eq. \eqref{eq:floatingCentroidal} and the joints dynamics Eq. \eqref{eq:jointsCentroidal}. We define $h:={C}(q, {\nu}){\nu} + {G}(q)  \in \mathbb{R}^{n+6}$ and its partition $h = (h_b,h_s), h_b \in \mathbb{R}^6,  h_s \in \mathbb{R}^n$. $f:=(f_1,\cdots,f_{n_c}) \in \mathbb{R}^{6n_c} $ are the set of contact forces -- i.e. Lagrange multipliers -- making Eq.~\eqref{eq:constraints_acc} satisfied.
 
\subsection{Control with no Friction Exploitation}

We recall here the momentum-based control strategy for balancing implemented on the iCub humanoid robot~\cite{nava2016,pucciTuning2016}. More specifically, the control objective is the achievement of a desired robot momentum and the stability of the associated zero dynamics. 

\subsubsection{Momentum control}

recall that the rate-of-change of the robot momentum equals the net external wrench acting on the robot, which in the present case reduces to the contact wrenches $f$ plus the gravity wrench. In view of Eq. \eqref{centrTrans}--\eqref{eq:floatingCentroidal}, the rate-of-change of the robot momentum is given by:
\begin{IEEEeqnarray}{RCCCL}
      \label{hDot}
      \frac{\dif }{\dif t}(M_b {\text{v}_\mathcal{B}}) &=& \dot{H}(f) &=& J_b^{\top}f - mge_3.
\end{IEEEeqnarray}
Let $H^d \in \mathbb{R}^6 $ denote the desired robot momentum, and  $\tilde{H} = H - H^d$  the momentum error. Assuming  that the contact wrenches $f$ can be chosen at will, then we choose $f :=  f^*$ such that \cite{nava2016}:
\begin{IEEEeqnarray}{RCL}
      \label{hDotDes}
      \dot{H}(f^*) &=& 
      \dot{H}^* := \dot{H}^d - K_p \tilde{H} - K_i I_{\tilde{H}}    
      \IEEEeqnarraynumspace  \IEEEyessubnumber \label{eq:dotH_f}  \\
      \dot{I}_{\tilde{H}} &=&
                     \begin{bmatrix} 
                     {J}_{G}^L(s) \\ 
                     {J}_{G}^{\omega}(s^d)
                     \end{bmatrix}(\dot{s}-\dot{s}^d) \IEEEyessubnumber  \label{eq:IhTilde} 
\end{IEEEeqnarray}
with $K_p, K_i {\in} \mathbb{R}^{6\times 6}$ two symmetric and positive definite matrices, $I_{\tilde{H}}$ is an approximation of the momentum error integral, and ${J}_{G}^L, {J}_{G}^\omega$ are related to the so-called \emph{centroidal momentum matrix} $\bar{J}_G(s) \in \mathbb{R}^{6\times n}$, i.e. the matrix such that $H = \bar{J}_G(s)\dot{s}$, as follows \cite{Orin2013}:
\begin{IEEEeqnarray}{RCL}
      \label{eqn:jacobian} 
      \bar{J}_G(s) &{:=}& - M_bJ^{\dagger}_bJ_j =  \begin{bmatrix} {J}_{G}^L(s) \\ {J}_{G}^{\omega}(s)\end{bmatrix}. \IEEEeqnarraynumspace 
\end{IEEEeqnarray} 

\noindent If $n_c > 1$, there are infinite contact wrenches $f^*$ that satisfy Eq.~\eqref{eq:dotH_f}. We parametrize the set of solutions to \eqref{eq:dotH_f} as:
\begin{equation}
       \label{eq:forces}
       f^* = f_1 + N_{b}f_0
\end{equation}
with $f_1 =  J_b^{\top\dagger} \left(\dot{H}^*+ mg e_3\right)$, $N_b \in \mathbb{R}^{6n_c \times 6n_c}$ a projector into the null space of $J_b^{\top}$, and $f_0\in \mathbb{R}^{6n_c}$ the wrench redundancy that does not influence $\dot{H}(f^*) = \dot{H}^*$.
To determine the control torques that instantaneously realize the contact wrenches given by \eqref{eq:forces}, we substitute the state accelerations from the dynamic equations \eqref{eq:system} into the constraints equations \eqref{eq:constraints_acc}, which yields:
    \begin{equation}
      \label{eq:torques}
      \tau^* = \Lambda^\dagger (JM^{-1}(h - J^\top f^*) - \dot{J}\nu) + N_\Lambda \tau_0
\end{equation}
with $\Lambda = {J_s}{M_s}^{-1} \in \mathbb{R}^{6n_c\times n}$,  $N_\Lambda \in \mathbb{R}^{n\times n}$ a projector onto the nullspace of $\Lambda$,  and $\tau_0 \in \mathbb{R}^n$  a free variable.

\subsubsection{Stability of the Zero Dynamics}
  
the stability of the zero dynamics is attempted by means of the so called \emph{postural task}, which exploits the free variable $\tau_0$ in \eqref{eq:torques}. A choice of the postural task that ensures the stability of the zero dynamics in case of one foot balancing is \cite{nava2016}:
\begin{IEEEeqnarray}{lCr}
      \label{posturalNew}
      \tau_0 &=& h_s - J_s^\top f + u_0
\end{IEEEeqnarray}
with $u_0 := -K^s_{p}N_\Lambda M_s(s-s^d) -K^s_{d}N_\Lambda M_s(\dot{s}-\dot{s}^d)$. $K^{s}_p, \ K^{s}_d \in \mathbb{R}^{n \times n}$ are two symmetric, positive definite matrices. An interesting property of the closed loop system~\eqref{eq:system}--\eqref{eq:forces}--\eqref{eq:torques}--\eqref{posturalNew} is that in view of the choice~\eqref{posturalNew} of the postural control, the closed loop joint space dynamics does not depend upon the wrench redundancy $f_0$, that can be chosen e.g. to minimize the joint torques $\tau^* = \tau^*(f_0)$.

In the language of Optimization Theory, we can rewrite the  control strategy as the following optimization problem:
\begin{IEEEeqnarray}{RCL}
	  \IEEEyesnumber
	  \label{inputTorquesSOT}
	  f^* &=& \text{argmin}_{f}  |\tau^*(f)|^2 \IEEEyessubnumber \label{inputTorquesSOTMinTau}  \\
		   &s.t.& \nonumber \\
		   &&Af \leq b \IEEEyessubnumber  \label{frictionCones} \\
		   &&\dot{H}(f) = \dot{H}^*  \IEEEyessubnumber \\
		   &&\tau^*(f) = \text{argmin}_{\tau}  |\tau - \tau_0(f)|^2 \IEEEyessubnumber	\label{optPost} 
    \\
		   	&& \quad s.t.  \nonumber \\
		   	&& \quad \quad \ \dot{J}(q,\nu)\nu + J(q)\dot{\nu} = 0
		    \IEEEyessubnumber 	\label{constraintsRigid} \\
	     	&& \quad \quad \ \dot{\nu} = M^{-1}(B\tau+J^\top f {-} h) \IEEEyessubnumber \\
		    && \quad \quad \ 	\tau_0 = 
		    h_s - J_s^\top f + u_0.		    \IEEEyessubnumber
\end{IEEEeqnarray}
The constraints~\eqref{frictionCones} ensure the satisfaction of friction cones, normal contact surface forces, and center-of-pressure constraints. The control torques are then given by $\tau {=} \tau^*(f^*)$. \\

\subsubsection{Joint Torques Control}

another control loop is responsible for stabilizing the actual joints torques $\tau$ towards the reference $\tau^*$. More specifically, we assume that any reasonable motor torques $\tau_m$ can be (almost) instantly achieved by means of a fast current control loop at the motor level. This control also compensates for the motor's back electromotive effect. Then, we choose $\tau_m := \tau_m^*$ as follows:
\begin{IEEEeqnarray}{LCL}
      \label{eq:motors-torques}
      \tau_m^* &=& K_v\dot{\theta} +K_c \text{sign}(\dot{\theta}) +\Gamma^{\top}(\tau^* -K_I\int_0^t{\tilde{\tau}} \ dt).
\end{IEEEeqnarray}
$K_I \in \mathbb{R}^{n \times n}$ is a symmetric and positive definite matrix and $\tilde{\tau} = \tau -\tau^*$. We substitute Eq. \eqref{eq:motors-torques} into the motors dynamics Eq. \eqref{eq:motorsDynamics} to obtain the following closed loop dynamics for the joint torques:
\begin{IEEEeqnarray}{LCL}
      \label{eq:closedLoopMotors}
      \tau = \tau^* -\Gamma^{-\top}I_m\ddot{\theta}  -K_I\int_0^t{\tilde{\tau}} \ dt.
\end{IEEEeqnarray}
We refer to Eq. \eqref{eq:motors-torques}--\eqref{eq:closedLoopMotors} as the \emph{inner} control loop. Note that we do not compensate for the term $I_m\ddot{\theta}$ through Eq. \eqref{eq:motors-torques} because its magnitude is usually negligible compared to that of the other terms of Eq. \eqref{eq:closedLoopMotors}.

%% file: tex/method.tex
\section{EXPLOITING FRICTION}
\label{sec:method}

The two-loops control architecture Eq. \eqref{inputTorquesSOT}--\eqref{eq:closedLoopMotors} is a common design for performing whole-body torque control of humanoid robots \cite{delPrete2016}. However, it compensates for the effect of joints viscous and Coulomb friction, thus removing also their inherent stabilizing property that may be beneficial for achieving high level control objectives. Furthermore, the friction compensation terms in Eq. \eqref{eq:motors-torques} may render the system sensitive to poor velocity measurements. In what follows, we design instead a control architecture that aims at increasing the robustness of the system w.r.t. poor velocity measurements, and also exploits friction to improve the tracking performances of both joints and momentum reference trajectories. 

\subsection{Reformulation of the System Dynamics}
 
Recall the kinematic relation Eq. \eqref{eq:jointMotorConstraint} between the joints position $s$ and motors position $\theta$, and rewrite the motors dynamics Eq. \eqref{eq:motorsDynamics} by substituting $\ddot{\theta}$, $\dot{\theta}$ with $\ddot{s}$ and $\dot{s}$: 
\begin{IEEEeqnarray}{LCL}
    \label{eq:motors-with-s}
    I_m\Gamma^{-1}\ddot{s} = \tau_m -K_f\Gamma^{-1}\dot{s} -\Gamma^{\top}\tau
\end{IEEEeqnarray}
where $K_f = K_f(\dot{s})$ is a diagonal matrix that collects the Coulomb and viscous friction coefficients. The elements along the diagonal of $K_f$ are all positive, and of the form: 
\begin{IEEEeqnarray}{LCL}
    \label{eq:KfCoeff}
    k_{f_{(i)}} = k_{v_{(i)}} + \frac{k_{c_{(i)}}}{|e_i^{\top}\Gamma^{-1}\dot{s}|+\epsilon}, \ \ i = 1 ... n,
\end{IEEEeqnarray}
with $e_i$ the canonical vector, consisting of all zeros but the $i$-th component that is equal to one, and the $\epsilon \in \mathbb{R}^+$, $\epsilon << 1$. The coefficients in Eq. \eqref{eq:KfCoeff} are obtained by rewriting the $\text{sign}$ function as $\text{sign}(\dot{\theta}) = \frac{\dot{\theta}}{|\dot{\theta}|}$. The parameter $\epsilon$ is a regularization term that avoids the coefficients to reach infinite values when $\dot{s} \xrightarrow[]{} 0$. A more detailed discussion on how to properly choose $\epsilon$ is postponed to future work.

We multiply Eq. \eqref{eq:motors-with-s} times $\Gamma^{-\top}$ and we sum joints and motors equations of motion Eq. \eqref{eq:jointsCentroidal}-\eqref{eq:motors-with-s} to get:
  \begin{IEEEeqnarray}{LCL}
    \label{eq:equiv-system-fixed}
    \overline{M}_s\ddot{s} & = & u -h_s + J_s^{\top}f -\overline{K}_f\dot{s},  \\
    \text{with:} \nonumber \\
    \overline{M}_s & = & M_s +\Gamma^{-\top}I_m\Gamma^{-1},\nonumber \\
    \overline{K}_f &=& \Gamma^{-\top}K_f\Gamma^{-1}, \ \overline{K}_f=\overline{K}_f^{\top} > 0, \nonumber \\
    u &=& \Gamma^{-\top}\tau_m. \nonumber
\end{IEEEeqnarray}
More specifically, the term $\Gamma^{-\top}I_m\Gamma^{-1}$ is the so-called \emph{motor reflected inertia} \cite{schaffer2007}. It accounts for the effect of the motors inertia on the joint space dynamics, and it has a pivotal role in improving the numerical stability of the control algorithm when the control design requires to invert the joint space mass matrix. In fact, in case of a humanoid robot the matrix $M_s$ is often \emph{ill-conditioned} because of the presence the of links with very different mass and inertia properties. The motor reflected inertia may be also interpreted as a \emph{physically consistent} regularization term that decreases the condition number of the mass matrix, defined as $\text{cond}(M_s) = \frac{\sigma^{max}}{\sigma^{min}}$, with $\sigma^i$ the singular values of $M_s$. However, it is not straightforward to prove that the regularized mass matrix $\overline{M}_s  =  M_s +\Gamma^{-\top}I_m\Gamma^{-1}$ is no more ill-conditioned. We verified numerically this result by comparing the condition numbers of $M_s$ and $\overline{M}_s$. Further investigations are postponed to future work.

Finally, the robot and motors dynamics Eq. \eqref{eq:systemFullCentroidal} can be rewritten as the following system:
\begin{IEEEeqnarray}{LCL}
    \label{eq:systemFullCentroidalNew}
    M_b\dot{\text{v}}_\mathcal{B} + h_b  & = &  J_b^{\top}f     \IEEEyessubnumber \label{eq:floatingCentroidalNew} \\
  \overline{M}_s\ddot{s} +h_s & = & u  + J_s^{\top}f -\overline{K}_f\dot{s}.            \IEEEyessubnumber \label{eq:jointsCentroidalNew} 
\end{IEEEeqnarray} 

Now, system \eqref{eq:systemFullCentroidalNew} may be stabilized using the variable $u$ as control input instead of the joint torques. 

\subsection{Control of a Fixed Base Robot} 

For a better understanding of the motivations behind our control approach, at first we assume that the robot base link is fixed on a pole, i.e. $(\dot{\text{v}}_\mathcal{B},{\text{v}}_\mathcal{B}) = (0,0)$, and no other links are in contact with the environment. Therefore the system dynamics Eq. \eqref{eq:systemFullCentroidalNew} reduces to:
\begin{IEEEeqnarray}{LCL}
    \label{eq:jointsNoF}
    \overline{M}_s\ddot{s} +h_s & = & u -\overline{K}_f\dot{s}, 
\end{IEEEeqnarray}
that is, the joints dynamics Eq. \eqref{eq:jointsCentroidalNew} with $f = 0$. The control objective is the stabilization of a desired joints trajectory $(s,\dot{s}) = (s^d,\dot{s}^d)$. We choose the input $u := u^*$ as:
\begin{IEEEeqnarray}{LCL}
    \label{eq:u-fixed}
    u^* = h_s +\overline{M}_s\ddot{s}^d -K_p^s\tilde{s} -K_d^s\dot{\tilde{s}} + \overline{K}_f\dot{s}^d
\end{IEEEeqnarray}
with $K_p^s, \ K_d^s$ two symmetric and positive definite matrices and $\tilde{s} = s-s^d$. Substituting Eq. \eqref{eq:u-fixed} into Eq. \eqref{eq:jointsNoF} yields to the following closed-loop dynamics:
\begin{IEEEeqnarray}{LCL}
    \label{eq:closed-loop-joints-fixed}
    \overline{M}_s\ddot{\tilde{s}}  + (K_d^s +\overline{K}_f) \dot{\tilde{s}} + K_p^s\tilde{s} = 0. 
\end{IEEEeqnarray}
In particular, being $\overline{K}_f$ symmetric and positive definite, the term $\overline{K}_f\dot{\tilde{s}}$ in Eq. \eqref{eq:closed-loop-joints-fixed} enforces the feedback term on the joints velocity error. This allows to exploit the joints friction for improving the convergence of the closed loop system dynamics to the reference trajectory. More specifically, the idea of exploiting the passive properties of the system dynamics in the control design is typical of the \emph{passivity-based} control approach \cite{li2012},\cite{schaffer2007}.

The advantage of applying the control law \eqref{eq:u-fixed} rather than, e.g. a classical feedback linearization with friction compensation technique, is that it may guarantee better robustness w.r.t. poor velocity measurements. On real robotic applications the velocity measurements are often obtained by means of numerical differentiation of the joint/motor positions measurements, and the estimated values can be noisy, or delayed in case a filtering technique is applied to the signal. In the control law Eq. \eqref{eq:u-fixed} the matrix $\overline{K}_f(\dot{s})$ is multiplied by the reference velocity $\dot{s}^d$ instead of the measured one, thus rendering the control algorithm less sensitive to the noise on the velocity measurements. Furthermore, in most control algorithms, the gain matrix that multiplies the joints velocity error $K_d^s$ is limited to relatively small values, because high values of $K_d^s$ may lead to numerical instability. The additional term $\overline{K}_f\dot{\tilde{s}}$ in the closed loop dynamics Eq. \eqref{eq:closed-loop-joints-fixed} contributes to increase the system damping without the need of modifying $K_d^s$, and it may improve the tracking performances of the controlled system. \\

\noindent \textbf{Remark:} the control law Eq. \eqref{eq:u-fixed} belongs to a family of controllers of the form: 
\begin{IEEEeqnarray}{LCL}
    \label{eq:u-family}
    u_f : = h_s +\overline{M}_s\ddot{s}^d -K_p^s\tilde{s} + \bar{u}(K,\dot{s},\dot{s}^d), \\
    \bar{u} = -K\dot{\tilde{s}} + \overline{K}_f\dot{s}, \nonumber \\
    K = K^{\top}, \ K > 0. \nonumber
\end{IEEEeqnarray}
Note that for any symmetric and positive definite matrix $K$, substituting $u_f$ from Eq. \eqref{eq:u-family} in the joint space dynamics Eq. \eqref{eq:jointsNoF} always guarantees stability and convergence of the closed loop system dynamics to the reference trajectory. Among all the possible choices of the gain $K$, we may be interested in finding the one that minimizes the sensitivity of $\bar{u}$ w.r.t. the joint velocities $\dot{s}$, i.e.:
\begin{IEEEeqnarray}{LCL}
    \label{eq:K-opt}
    K^* : &=& \text{argmin}_K |\frac{\delta(\bar{u})}{\delta \dot{s}}|^2 \\
    & s.t. \nonumber \\
    & & K = K^{\top}, \ K > 0. \nonumber
\end{IEEEeqnarray}
Assume that $\overline{K}_f$ does not depend on the joint velocities $\dot{s}$ (i.e., $\overline{K}_f$ accounts for the viscous friction only, while the Coulomb friction is compensated by the inner control loop). Then, the solution to problem \eqref{eq:K-opt} is given by $K^* = \overline{K}_f$, which leads to $\bar{u}^* = \overline{K}_f\dot{s}^d$. Finally, the corresponding control input $u_f^*$ is given by: $u_f^* = h_s +\overline{M}_s\ddot{s}^d -K_p^s\tilde{s} + \overline{K}_f\dot{s}^d$, 
that coincides with Eq. \eqref{eq:u-fixed} when we choose $K_d^s = 0_n$. In this sense, the control law Eq. \eqref{eq:u-fixed} with $K_d^s = 0_n$ can be seen as the one among the family of controllers $u_f$ that is less sensitive to the joints velocity, and it is therefore the most robust (among $u_f$) against poor velocity measurements. The extension of this theoretical framework in the more general case $\overline{K}_f = \overline{K}_f(\dot{s})$ will be addressed in future work.

\subsection{Control of a Floating Base Robot}
   
In what follows we propose a modification of the momentum-based control algorithm Eq. \eqref{eq:forces}--\eqref{posturalNew} that allows to exploit friction for improving the tracking of a desired momentum trajectory. As for Eq. \eqref{eq:system}, we compactly rewrite system \eqref{eq:systemFullCentroidalNew} as follows:
\begin{IEEEeqnarray}{LCL}
    \label{eq:system-reduced}
    {\overline{M}}\dot{{\nu}} + h & = &  J^{\top}f +Bu -B\overline{K}_f\dot{s},
\end{IEEEeqnarray} 
where we recall that $h = (h_b,h_s)$, $J = (J_b, J_s)$ and the selector matrix $B$ is of the form $B = (0_{n\times 6} , 1_n)^\top$. The mass matrix $\overline{M}$ is given by:
$\overline{M} = \begin{bmatrix} {M}_b & 0_{6\times n} \\ 0_{n\times 6} & \overline{M}_s \end{bmatrix}.$
The friction component $\overline{K}_f\dot{s}$ can be related to the contact wrenches $f$ by means of the contact constraint equations Eq. \eqref{eq:constraints_acc}. In particular, we substitute the state acceleration $\dot{\nu} = {\overline{M}}^{-1}( J^{\top}f -h +Bu -B\overline{K}_f\dot{s})$ obtained by inverting Eq. \eqref{eq:system-reduced} into the constraint \eqref{eq:constraints_acc}, which yields:
\begin{IEEEeqnarray}{LCL}
    \label{eq:forces-friction}
    J\overline{M}^{-1}(J^{\top}f -h +Bu -B\overline{K}_f\dot{s}) +\dot{J}\nu = 0.
\end{IEEEeqnarray}
Writing explicitly the contact wrenches from Eq. \eqref{eq:forces-friction} gives:
\begin{IEEEeqnarray}{LCL}
    \label{eq:forces-fm-fe}
    f = f_m + D\overline{K}_f\dot{s},
\end{IEEEeqnarray}
where we define $f_m, D$ as:
\begin{IEEEeqnarray}{LCL}
    \label{eq:f_m-D}
    f_m &=& (J\overline{M}^{-1}J^{\top})^{-1}( J\overline{M}^{-1}(h -Bu) -\dot{J}\nu) \\
    D   &=& (J\overline{M}^{-1}J^{\top})^{-1}J\overline{M}^{-1}B. \nonumber
\end{IEEEeqnarray} 
Recall that the rate-of-change of the robot momentum Eq. \eqref{hDot} equals the net external wrench acting on the robot, and substitute the contact wrenches $f$ obtained from Eq. \eqref{eq:forces-fm-fe} into the momentum dynamics Eq. \eqref{hDot}:
\begin{IEEEeqnarray}{LCL}
    \label{eq:forces-H-modified}
    \dot{H}(f) = J_b^{\top}f_m + J_b^{\top}D\overline{K}_f\dot{s} - mge_3.
\end{IEEEeqnarray} 
In order to come up with a formulation similar to that of Eq. \eqref{eq:equiv-system-fixed}, we split the joints velocity $\dot{s}$ into two components:
\begin{IEEEeqnarray}{LCL}
    \label{eq:dot_s_partititon}
    \dot{s} &=& -D^{\top}J_b\bar{J}_G\dot{s}+ (1_n +D^{\top}J_b\bar{J}_G)\dot{s},
\end{IEEEeqnarray}
where $\bar{J}_G$ is the \emph{centroidal momentum matrix} as defined in Eq. \eqref{eqn:jacobian}. In particular, $\bar{J}_G$ is the mapping between the joint velocities $\dot{s}$ and the momentum, i.e. $H = \bar{J}_G\dot{s}$. Hence, Eq. \eqref{eq:dot_s_partititon} can be rewritten as: $\dot{s} = -D^{\top}J_bH+ (1_n +D^{\top}J_b\bar{J}_G)\dot{s}$.

Assuming that $f_m$ in Eq. \eqref{eq:forces-H-modified} can be chosen at will, we choose $f_m^*$ as:
\begin{IEEEeqnarray}{LCL}
    \label{eq:newHdes}
    f_m^* = f_{m1} + N_{b}f_{m0}, \\
    \text{with} \nonumber \\
    f_{m1} = J_b^{\top\dagger} \left(\dot{H}^* -J_b^{\top}D\overline{K}_f(1_n +D^{\top}J_b\bar{J}_G)\dot{s} +mg e_3 \right), \nonumber \\
    \dot{H}^* = \dot{H}^d - K_p \tilde{H} - K_i I_{\tilde{H}} +TH^d, \nonumber 
\end{IEEEeqnarray}
where we recall that $N_b \in \mathbb{R}^{6n_c \times 6n_c}$ is a projector into the null space of $J_b^{\top}$, and $f_{m0}\in \mathbb{R}^{6n_c}$ is a free variable. $T = J_b^{\top}D\overline{K}_fD^{\top}J_b \in \mathbb{R}^{6 \times 6}$ is a symmetric and positive definite matrix. If $f_m^*$ is chosen as in Eq. \eqref{eq:newHdes}, the closed loop momentum dynamics remains: $\dot{\tilde{H}} + (K_p + T)\tilde{H} + K_i I_{\tilde{H}} = 0$. Note that the same considerations about robustness and tracking performances done for the fixed robot closed loop dynamics Eq. \eqref{eq:closed-loop-joints-fixed} can be applied also to the closed loop momentum dynamics.

To determine the control input $u^*$ that instantaneously realize $f_m^*$, we make use of Eq. \eqref{eq:f_m-D}, which yields:
    \begin{equation}
      \label{eq:uFloatingBase}
      u^* = \overline{\Lambda}^\dagger (J\overline{M}^{-1}(h - J^\top f_m^*) - \dot{J}\nu) + \overline{N}_\Lambda u_{null}
\end{equation}
with $\overline{\Lambda} = {J_s}{\overline{M}_s}^{-1} \in \mathbb{R}^{6n_c\times n}$,  $\overline{N}_\Lambda \in \mathbb{R}^{n\times n}$ a projector onto the nullspace of $\overline{\Lambda}$, and $u_{null} \in \mathbb{R}^n$  a free variable. Following the control law Eq. \eqref{eq:u-fixed} developed for stabilizing the joints dynamics of a fixed base robot, and recalling the choice of the joint torques redundancy $\tau_0$ as in Eq. \eqref{posturalNew}, we design the postural task $u_{null}$ as follows:
\begin{IEEEeqnarray}{lCr}
      \label{posturalFric}
      u_{null} &=& h_s - J_s^\top f +\overline{K}_f\dot{s}^d + u_0,
\end{IEEEeqnarray}
with $u_0 := -K^s_{p}\overline{N}_\Lambda \overline{M}_s\tilde{s} -K^s_{d}\overline{N}_\Lambda \overline{M}_s\dot{\tilde{s}}$. The additional term $\overline{K}_f\dot{s}^d$ may help to improve also the tracking of the postural task references. Asymptotic stability of the closed loop system Eq. \eqref{eq:system-reduced}--\eqref{eq:newHdes}--\eqref{eq:uFloatingBase}--\eqref{posturalFric} and the effectiveness of the control algorithm in improving the momentum tracking are verified experimentally.

  \begin{figure}[ht]
    \centering
	\def\svgwidth{0.8\linewidth}{\input{"iCub_Pole_JointErrors.pdf_tex"}}
	 \label{fig:pole_jointErrors}
	\caption{Norm of joint position error with the fixed robot. The control law that exploits friction (EF) shows better tracking performances.}
  \end{figure}
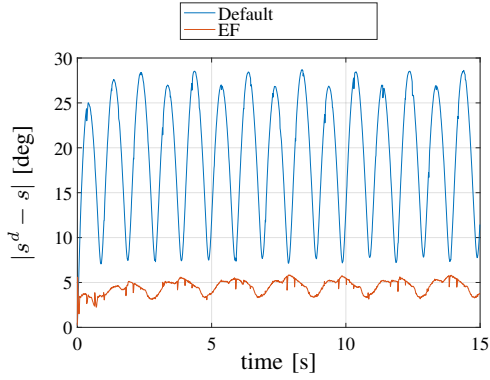
   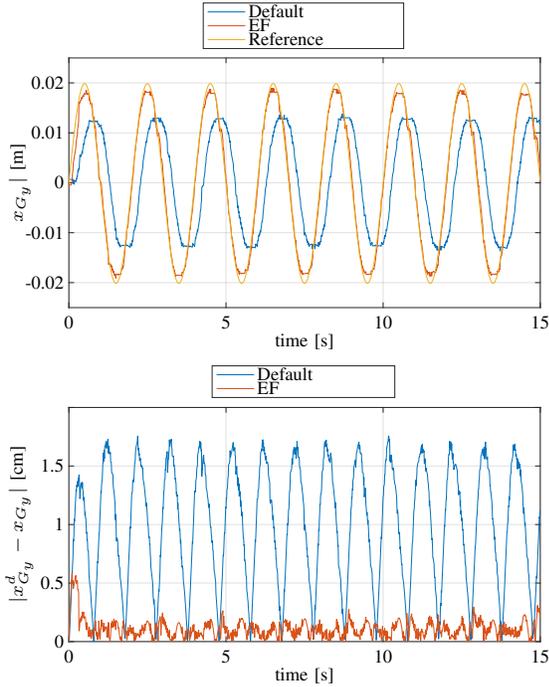
\begin{figure}[ht]
    \centering
    \label{fig:LR_comTracking}
	\def\svgwidth{\linewidth}{\input{"iCub_LR_CoM_tracking.pdf_tex"}}
	\caption{The upper figure shows the CoM reference trajectory versus the CoM measured position, while the lower figure is the error norm of the CoM position. The control law that exploits friction (EF) shows better tracking performances.}
  \end{figure}

\subsection{Inner Control Loop}

The inner control loop Eq. \eqref{eq:motors-torques}--\eqref{eq:closedLoopMotors} is modified in order stabilize the input $u$ towards the reference $u^*$. More specifically, recall that $u = \Gamma^{-\top}\tau_m$. Then, we choose $\tau_m^*$ as:
\begin{IEEEeqnarray}{LCL}
    \label{eq:tau_m-friction}
    \tau_m^* = \Gamma^{\top}(u^* -K_I\int_0^t{(u-u^*)} \ dt),
\end{IEEEeqnarray}
that yields to the following closed loop dynamics for the control input: $u  = u^* -K_I\int_0^t{(u-u^*)} \ dt$.

%% file: tex/iCub_Pole_JointErrors.pdf_tex
\begingroup%
  \makeatletter%
  \providecommand\color[2][]{%
    \errmessage{(Inkscape) Color is used for the text in Inkscape, but the package 'color.sty' is not loaded}%
    \renewcommand\color[2][]{}%
  }%
  \providecommand\transparent[1]{%
    \errmessage{(Inkscape) Transparency is used (non-zero) for the text in Inkscape, but the package 'transparent.sty' is not loaded}%
    \renewcommand\transparent[1]{}%
  }%
  \providecommand\rotatebox[2]{#2}%
  \ifx\svgwidth\undefined%
    \setlength{\unitlength}{896bp}%
    \ifx\svgscale\undefined%
      \relax%
    \else%
      \setlength{\unitlength}{\unitlength * \real{\svgscale}}%
    \fi%
  \else%
    \setlength{\unitlength}{\svgwidth}%
  \fi%
  \global\let\svgwidth\undefined%
  \global\let\svgscale\undefined%
  \makeatother%
  \begin{picture}(1,0.75)%
    \put(0,0){\includegraphics[width=\unitlength,page=1]{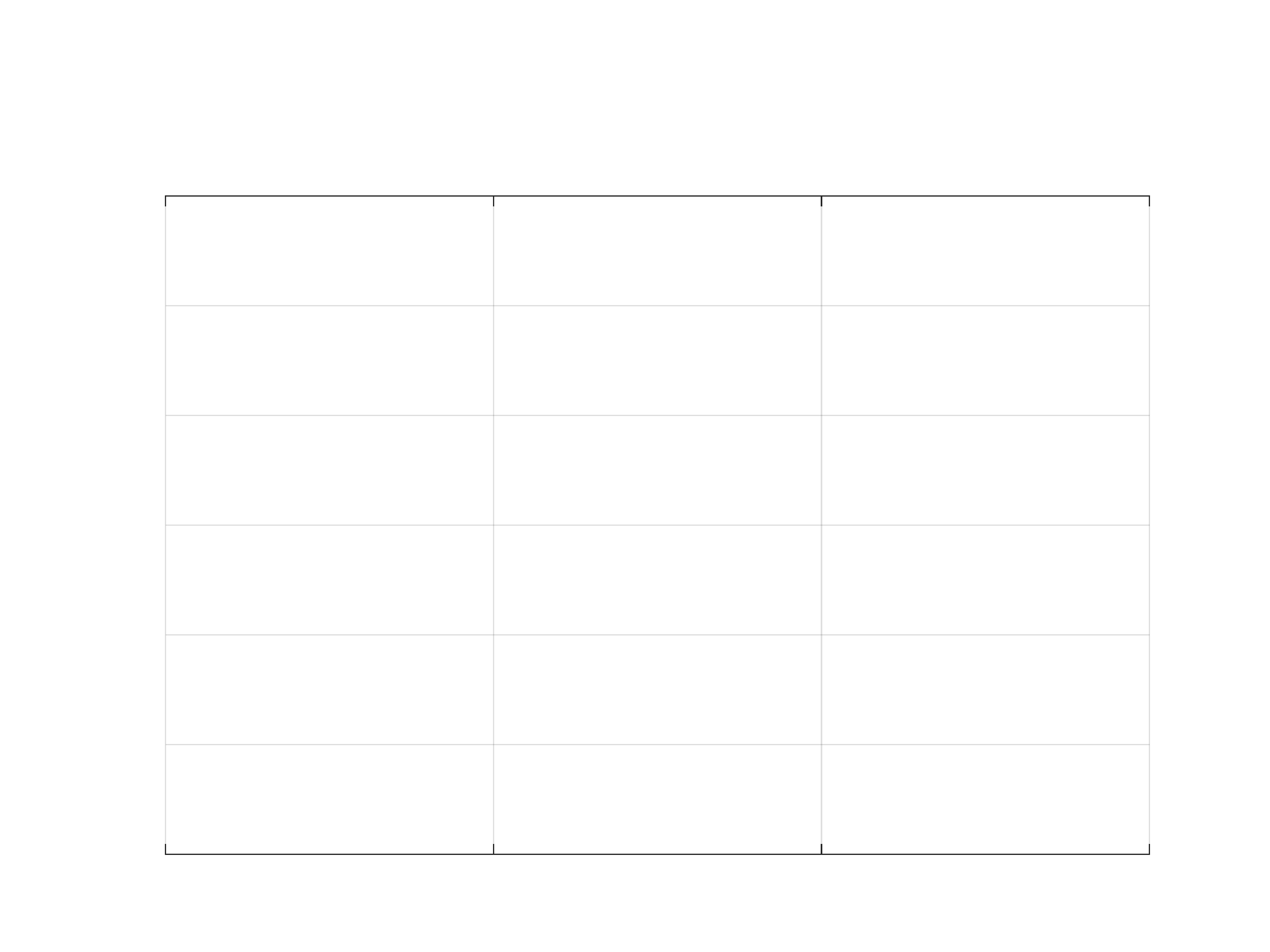}}%
    \put(0.13035714,0.03254464){\makebox(0,0)[b]{\smash{\scriptsize{0}}}}%
    \put(0.38869045,0.03254464){\makebox(0,0)[b]{\smash{\scriptsize{5}}}}%
    \put(0.64702384,0.03254464){\makebox(0,0)[b]{\smash{\scriptsize{10}}}}%
    \put(0.90535714,0.03254464){\makebox(0,0)[b]{\smash{\scriptsize{15}}}}%
    \put(0.5178575,-0.00357143){\makebox(0,0)[b]{\smash{\small{time [s]}}}}%
    \put(0,0){\includegraphics[width=\unitlength,page=2]{iCub_Pole_JointErrors.pdf}}%
    \put(0.12111607,0.06366071){\makebox(0,0)[rb]{\smash{\scriptsize{0}}}}%
    \put(0.12111607,0.15011902){\makebox(0,0)[rb]{\smash{\scriptsize{5}}}}%
    \put(0.12111607,0.23657741){\makebox(0,0)[rb]{\smash{\scriptsize{10}}}}%
    \put(0.12111607,0.32303571){\makebox(0,0)[rb]{\smash{\scriptsize{15}}}}%
    \put(0.12111607,0.40949402){\makebox(0,0)[rb]{\smash{\scriptsize{20}}}}%
    \put(0.12111607,0.49595241){\makebox(0,0)[rb]{\smash{\scriptsize{25}}}}%
    \put(0.12111607,0.58241071){\makebox(0,0)[rb]{\smash{\scriptsize{30}}}}%
    \put(0.03723214,0.33616098){\rotatebox{90}{\makebox(0,0)[b]{\smash{\small{$|s^d - s|$ [deg]}}}}}%
    \put(0,0){\includegraphics[width=\unitlength,page=3]{iCub_Pole_JointErrors.pdf}}%
    \put(0.40125357,0.66946429){\makebox(0,0)[lb]{\smash{\scriptsize{Default}}}}%
    \put(0,0){\includegraphics[width=\unitlength,page=4]{iCub_Pole_JointErrors.pdf}}%
    \put(0.40125357,0.63732143){\makebox(0,0)[lb]{\smash{\scriptsize{EF}}}}%
    \put(0,0){\includegraphics[width=\unitlength,page=5]{iCub_Pole_JointErrors.pdf}}%
  \end{picture}%
\endgroup%

%% file: tex/iCub_LR_CoM_tracking.pdf_tex
\begingroup%
  \makeatletter%
  \providecommand\color[2][]{%
    \errmessage{(Inkscape) Color is used for the text in Inkscape, but the package 'color.sty' is not loaded}%
    \renewcommand\color[2][]{}%
  }%
  \providecommand\transparent[1]{%
    \errmessage{(Inkscape) Transparency is used (non-zero) for the text in Inkscape, but the package 'transparent.sty' is not loaded}%
    \renewcommand\transparent[1]{}%
  }%
  \providecommand\rotatebox[2]{#2}%
  \ifx\svgwidth\undefined%
    \setlength{\unitlength}{936bp}%
    \ifx\svgscale\undefined%
      \relax%
    \else%
      \setlength{\unitlength}{\unitlength * \real{\svgscale}}%
    \fi%
  \else%
    \setlength{\unitlength}{\svgwidth}%
  \fi%
  \global\let\svgwidth\undefined%
  \global\let\svgscale\undefined%
  \makeatother%
  \begin{picture}(1,1.12820513)%
    \put(0,0){\includegraphics[width=\unitlength,page=1]{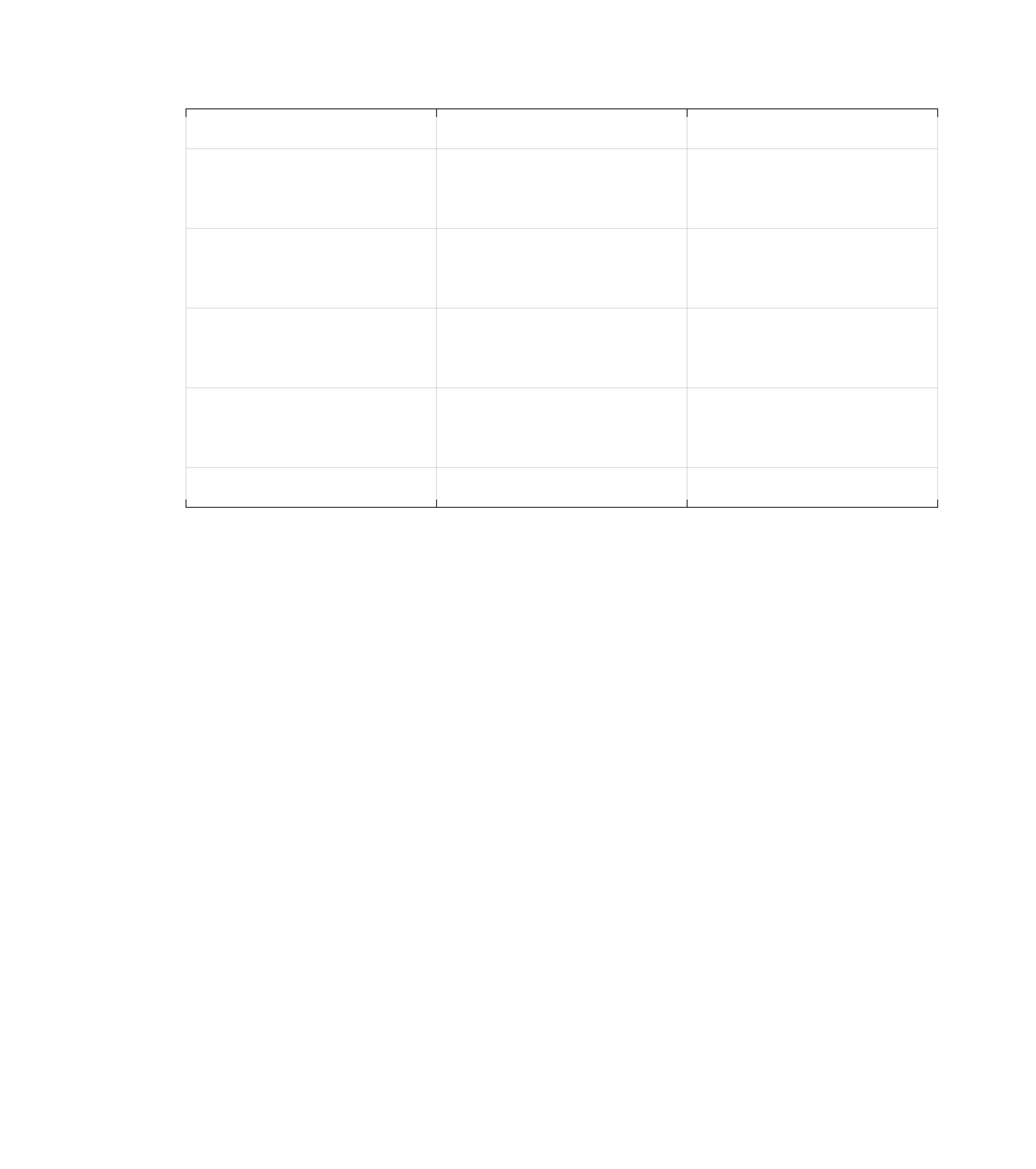}}%
    \put(0.17948718,0.60564103){\makebox(0,0)[b]{\smash{\scriptsize{0}}}}%
    \put(0.42136752,0.60564103){\makebox(0,0)[b]{\smash{\scriptsize{5}}}}%
    \put(0.66324786,0.60564103){\makebox(0,0)[b]{\smash{\scriptsize{10}}}}%
    \put(0.90512821,0.60564103){\makebox(0,0)[b]{\smash{\scriptsize{15}}}}%
    \put(0.54230803,0.57824786){\makebox(0,0)[b]{\smash{\scriptsize{time [s]}}}}%
    \put(0,0){\includegraphics[width=\unitlength,page=2]{iCub_LR_CoM_tracking.pdf}}%
    \put(0.17179487,0.66794872){\makebox(0,0)[rb]{\smash{\scriptsize{-0.02}}}}%
    \put(0.17179487,0.74487179){\makebox(0,0)[rb]{\smash{\scriptsize{-0.01}}}}%
    \put(0.17179487,0.82179487){\makebox(0,0)[rb]{\smash{\scriptsize{0}}}}%
    \put(0.17179487,0.89871795){\makebox(0,0)[rb]{\smash{\scriptsize{0.01}}}}%
    \put(0.17179487,0.97564103){\makebox(0,0)[rb]{\smash{\scriptsize{0.02}}}}%
    \put(0.1084188,0.8307694){\rotatebox{90}{\makebox(0,0)[b]{\smash{\scriptsize{$x_{G_y} | $ [m]}}}}}%
    \put(0,0){\includegraphics[width=\unitlength,page=3]{iCub_LR_CoM_tracking.pdf}}%
    \put(0.45607949,1.08564103){\makebox(0,0)[lb]{\smash{\scriptsize{Default}}}}%
    \put(0,0){\includegraphics[width=\unitlength,page=4]{iCub_LR_CoM_tracking.pdf}}%
    \put(0.45607949,1.0642735){\makebox(0,0)[lb]{\smash{\scriptsize{EF}}}}%
    \put(0,0){\includegraphics[width=\unitlength,page=5]{iCub_LR_CoM_tracking.pdf}}%
    \put(0.45607949,1.04290598){\makebox(0,0)[lb]{\smash{\scriptsize{Reference}}}}%
    \put(0,0){\includegraphics[width=\unitlength,page=6]{iCub_LR_CoM_tracking.pdf}}%
    \put(0.17948718,0.09111111){\makebox(0,0)[b]{\smash{\scriptsize{0}}}}%
    \put(0.42136752,0.09111111){\makebox(0,0)[b]{\smash{\scriptsize{5}}}}%
    \put(0.66324786,0.09111111){\makebox(0,0)[b]{\smash{\scriptsize{10}}}}%
    \put(0.90512821,0.09111111){\makebox(0,0)[b]{\smash{\scriptsize{15}}}}%
    \put(0.54230803,0.06371795){\makebox(0,0)[b]{\smash{\scriptsize{time [s]}}}}%
    \put(0,0){\includegraphics[width=\unitlength,page=7]{iCub_LR_CoM_tracking.pdf}}%
    \put(0.17179487,0.11495726){\makebox(0,0)[rb]{\smash{\scriptsize{0}}}}%
    \put(0.17179487,0.20512821){\makebox(0,0)[rb]{\smash{\scriptsize{0.5}}}}%
    \put(0.17179487,0.29529915){\makebox(0,0)[rb]{\smash{\scriptsize{1}}}}%
    \put(0.17179487,0.38547009){\makebox(0,0)[rb]{\smash{\scriptsize{1.5}}}}%
    \put(0.17179487,0.47564103){\makebox(0,0)[rb]{\smash{}}}%
    \put(0.1084188,0.30427368){\rotatebox{90}{\makebox(0,0)[b]{\smash{\scriptsize{$|x_{G_y}^d - x_{G_y}|$ [cm]}}}}}%
    \put(0,0){\includegraphics[width=\unitlength,page=8]{iCub_LR_CoM_tracking.pdf}}%
    \put(0.4697641,0.52666667){\makebox(0,0)[lb]{\smash{\scriptsize{Default}}}}%
    \put(0,0){\includegraphics[width=\unitlength,page=9]{iCub_LR_CoM_tracking.pdf}}%
    \put(0.4697641,0.50529915){\makebox(0,0)[lb]{\smash{\scriptsize{EF}}}}%
    \put(0,0){\includegraphics[width=\unitlength,page=10]{iCub_LR_CoM_tracking.pdf}}%
  \end{picture}%
\endgroup%

%% file: tex/results.tex
\section{EXPERIMENTAL RESULTS}
\label{sec:results}

  \begin{figure}[!ht]
    \centering
    \label{fig:LR_errHlin}
	\def\svgwidth{\linewidth}{\input{"iCub_LR_errH_lin.pdf_tex"}}
	\caption{Linear momentum tracking error while balancing. The control law that exploits friction (EF) shows better tracking performances.}
\end{figure}
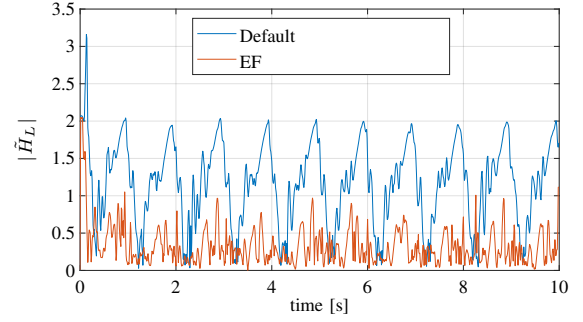

We tested the control algorithms presented in Sec. \ref{sec:method} on the iCub humanoid robot~\cite{Metta20101125}. For the purpose of this paper, iCub is endowed with $23$ degrees of freedom. The inner control loop runs at $1\mathrm{kHz}$, while the balancing controller runs at $100\mathrm{Hz}$. During all the experiments we only considered the effect of the viscous friction in the harmonic drive gearboxes, that on iCub gives the major contribution to friction effects while the robot is moving. This implies $k_{c_{(i)}} = 0 \ \forall i$.

\subsection{Joints Tracking on a Fixed Based Robot}

The first experimental setup is carried out with the robot \emph{pelvis} fixed on a pole. A sinusoidal reference trajectory of amplitude $15 [deg]$ and frequency $0.5 Hz$ is applied to each controlled joint. We evaluated the performances of the control algorithm Eq. \eqref{eq:u-fixed} and of a standard \emph{computed torque} control that compensates for the joints friction in the inner control loop. Fine tuning of the control gains has been performed in order to achieve the best possible tracking performances with both controllers. Figure (1) shows the norm of the joints position errors while executing the task. Despite this is not a proper performances comparison between the two controllers, it is possible to observe that the tracking performances of the default controller are limited by the small range of derivative gains that can be chosen without affecting the system stability. The control law Eq. \eqref{eq:u-fixed} (EF) allows instead to achieve better tracking performances without the need of increasing the derivative gains. A video showing the experiment is attached to the paper.

\subsection{Momentum Tracking on a Floating Based Robot}
   
The second experiment is carried out with the robot balancing on its feet. The robot moves its CoM from the left to the right foot, following a sinusoidal trajectory of amplitude $4 [cm]$ and frequency $0.5 Hz$. The center of mass trajectory can be tracked by controlling the robot's linear momentum dynamics. We evaluated the performances of the control  laws\eqref{eq:system}--\eqref{eq:forces}--\eqref{eq:torques}--\eqref{posturalNew}  and \eqref{eq:system-reduced}--\eqref{eq:newHdes}--\eqref{eq:uFloatingBase}--\eqref{posturalFric}. Figure (2) represents both the CoM error norm and the reference CoM trajectory signal versus the measured CoM position. Figure (3) represents the linear momentum error norm during left and right movements. As for the fixed base robot experiment, the tracking performance of the default controller is affected by the limited choice of the gain that multiply the momentum error $\tilde{H}$. The control law that exploits friction shows instead better tracking performances. A video of this second experiment is also presented. 

\subsection{The Contribution of Motors Reflected Inertia}

For the humanoid robot iCub the condition number of the joint space mass matrix is $\text{cond}(M_s) \approx 20000$. With the addition of motors reflected inertia as explained in Sec. \ref{sec:method} A, the condition number of the mass matrix decreased to $\text{cond}(\overline{M}_s) \approx 800$. The increased numerical stability of the control algorithm Eq. \eqref{eq:system-reduced}--\eqref{eq:newHdes}--\eqref{eq:uFloatingBase}--\eqref{posturalFric} allowed to perform very fast dynamic movements while balancing, that are extremely difficult to achieve with the default controller Eq. \eqref{eq:system}--\eqref{eq:forces}--\eqref{eq:torques}--\eqref{posturalNew}. A video of the robot performing very fast dynamic movements is attached to the paper.

%% file: tex/iCub_LR_errH_lin.pdf_tex
\begingroup%
  \makeatletter%
  \providecommand\color[2][]{%
    \errmessage{(Inkscape) Color is used for the text in Inkscape, but the package 'color.sty' is not loaded}%
    \renewcommand\color[2][]{}%
  }%
  \providecommand\transparent[1]{%
    \errmessage{(Inkscape) Transparency is used (non-zero) for the text in Inkscape, but the package 'transparent.sty' is not loaded}%
    \renewcommand\transparent[1]{}%
  }%
  \providecommand\rotatebox[2]{#2}%
  \ifx\svgwidth\undefined%
    \setlength{\unitlength}{896bp}%
    \ifx\svgscale\undefined%
      \relax%
    \else%
      \setlength{\unitlength}{\unitlength * \real{\svgscale}}%
    \fi%
  \else%
    \setlength{\unitlength}{\svgwidth}%
  \fi%
  \global\let\svgwidth\undefined%
  \global\let\svgscale\undefined%
  \makeatother%
  \begin{picture}(1,0.55)%
    \put(0,0){\includegraphics[width=\unitlength,page=1]{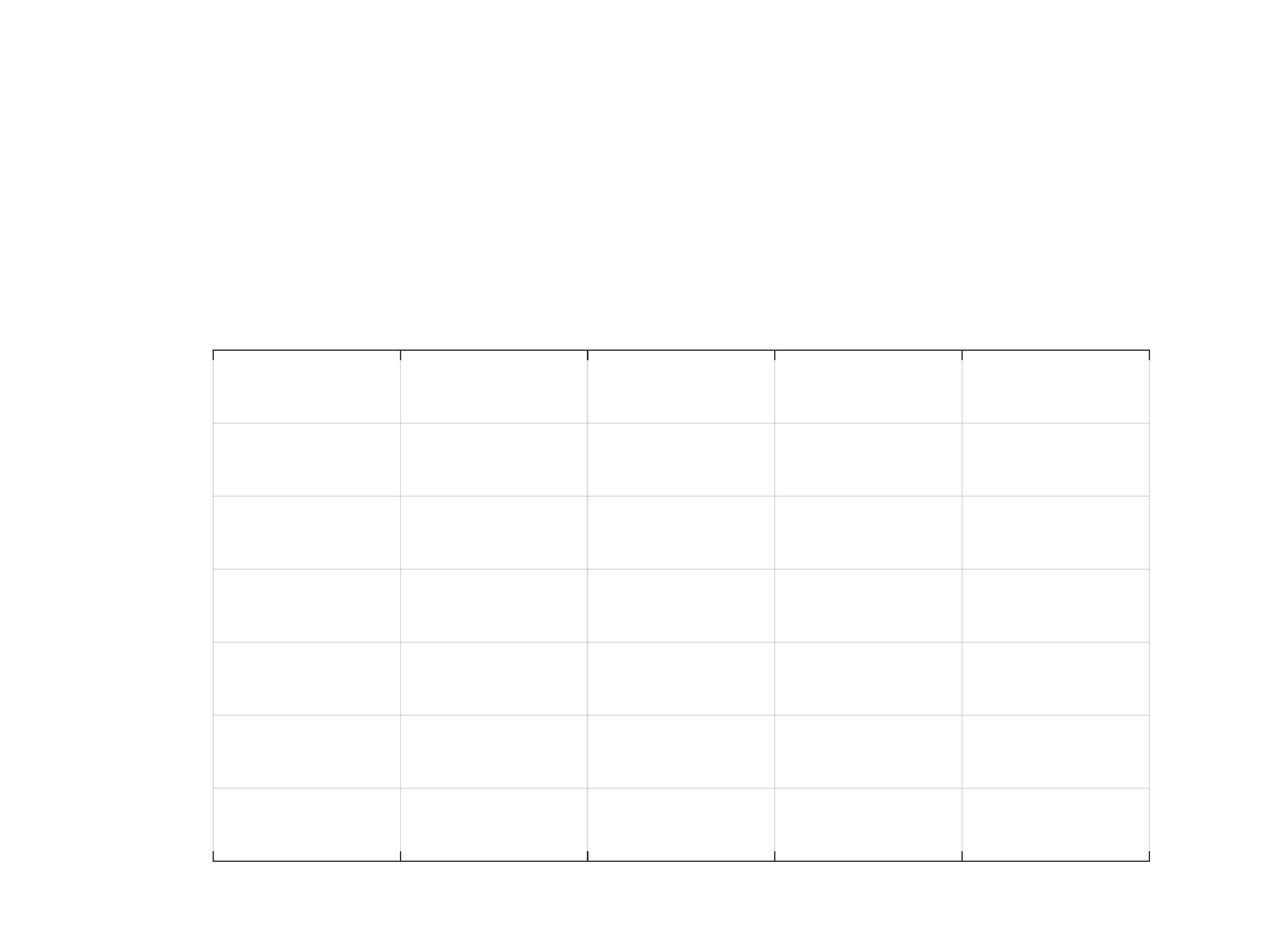}}%
    \put(0.16785714,0.03714286){\makebox(0,0)[b]{\smash{\scriptsize{0}}}}%
    \put(0.31535714,0.03714286){\makebox(0,0)[b]{\smash{\scriptsize{2}}}}%
    \put(0.46285714,0.03714286){\makebox(0,0)[b]{\smash{\scriptsize{4}}}}%
    \put(0.61035714,0.03714286){\makebox(0,0)[b]{\smash{\scriptsize{6}}}}%
    \put(0.75785714,0.03714286){\makebox(0,0)[b]{\smash{\scriptsize{8}}}}%
    \put(0.90535714,0.03714286){\makebox(0,0)[b]{\smash{\scriptsize{10}}}}%
    \put(0.5366075,0.00852679){\makebox(0,0)[b]{\smash{\scriptsize{time [s]}}}}%
    \put(0,0){\includegraphics[width=\unitlength,page=2]{iCub_LR_errH_lin.pdf}}%
    \put(0.15982143,0.06205357){\makebox(0,0)[rb]{\smash{\scriptsize{0}}}}%
    \put(0.15982143,0.11957911){\makebox(0,0)[rb]{\smash{\scriptsize{0.5}}}}%
    \put(0.15982143,0.17710455){\makebox(0,0)[rb]{\smash{\scriptsize{1}}}}%
    \put(0.15982143,0.23463009){\makebox(0,0)[rb]{\smash{\scriptsize{1.5}}}}%
    \put(0.15982143,0.29215562){\makebox(0,0)[rb]{\smash{\scriptsize{2}}}}%
    \put(0.15982143,0.34968116){\makebox(0,0)[rb]{\smash{\scriptsize{2.5}}}}%
    \put(0.15982143,0.40720661){\makebox(0,0)[rb]{\smash{\scriptsize{3}}}}%
    \put(0.15982143,0.46473214){\makebox(0,0)[rb]{\smash{\scriptsize{3.5}}}}%
    \put(0.09361607,0.27276804){\rotatebox{90}{\makebox(0,0)[b]{\smash{\scriptsize{$|\tilde{H}_{L}|$}}}}}%
    \put(0,0){\includegraphics[width=\unitlength,page=3]{iCub_LR_errH_lin.pdf}}%
    \put(0.41375723,0.42312616){\makebox(0,0)[lb]{\smash{\scriptsize{Default}}}}%
    \put(0,0){\includegraphics[width=\unitlength,page=4]{iCub_LR_errH_lin.pdf}}%
    \put(0.41375723,0.38365955){\makebox(0,0)[lb]{\smash{\scriptsize{EF}}}}%
    \put(0,0){\includegraphics[width=\unitlength,page=5]{iCub_LR_errH_lin.pdf}}%
  \end{picture}%
\endgroup%

%% file: tex/conclusions.tex
\section{CONCLUSIONS}
\label{sec:conclusions}

Classical algorithms for whole-body torque control of humanoid robots compensate for the effect of the joints viscous and Coulomb friction. This paper proposed instead a torque control framework that allows to exploit the inherent stabilizing nature of the joints friction. We first developed a control algorithm for fixed base robots that exploits friction for improving the tracking of a desired joint space trajectory, and that can also ensure better robustness of the controlled system w.r.t. noisy velocity measurements. Then, we extended our formulation to the control of a floating base robot. In particular, the joints friction was exploited to improve the tracking of a desired momentum trajectory. Experimental results on the iCub humanoid robot show the effectiveness of the proposed approach. Future work may focus on a better understanding of the theoretical framework and on the design of new experiments for validating the control algorithm in different and more challenging scenarios.